\documentclass[journal]{IEEEtran}

\usepackage{comment}
\usepackage{epsfig}
\usepackage{graphicx}
\usepackage{amsmath}
\usepackage{amssymb}
\usepackage{comment}
\usepackage{multirow}

\usepackage{hyperref}
\hypersetup{hidelinks}

\usepackage{booktabs}
\usepackage{array, caption, threeparttable}
\usepackage{caption}
\usepackage{subfigure}
\usepackage{algorithm}
\usepackage{listings}
\usepackage{color}
\usepackage{mathtools}

\usepackage{todonotes}
\usepackage{gensymb}

\DeclareUnicodeCharacter{2212}{-}
\DeclareUnicodeCharacter{00A0}{ }

\begin{document}

\title{Visual Prompt Flexible-Modal Face Anti-Spoofing}

\author{Zitong Yu, Rizhao Cai, Yawen Cui, Ajian Liu and Changsheng Chen

\thanks{Manuscript received July 2023. This work was supported by the Chinese National Natural Science Foundation Projects $\#$62072313. The first two authors contribute equally. Corresponding author: Changsheng Chen. }

\thanks{Z. Yu is with the Great Bay University, China (email: zitong.yu@ieee.org). }

\thanks{Rizhao Cai is with the ROSE Lab, School of EEE, Nanyang Technological University (email: rzcai@ntu.edu.sg).}

\thanks{Y. Cui is with the University of Oulu, Finland (email: yawen.cui@oulu.fi). }

\thanks{A. Liu is with the National Laboratory of Pattern Recognition, Institute of Automation, Chinese Academy of Sciences, and University of Chinese Academy of Sciences, Beijing 100190, China. (email: ajianliu92@gmail.com).}

\thanks{Changsheng Chen is with the Shenzhen University, China (email: cschen@szu.edu.cn).}
}
%\thanks{}}
%; revised August 26, 2015.

\markboth{IEEE TRANSACTIONS ON MULTIMEDIA}%
{Shell \MakeLowercase{\textit{et al.}}: Bare Demo of IEEEtran.cls for IEEE Journals}

\maketitle

\begin{abstract}
Recently, vision transformer based multimodal learning methods have been proposed to improve the robustness of face anti-spoofing (FAS) systems. However, multimodal face data collected from the real world is often imperfect due to missing modalities from various imaging sensors. Recently, flexible-modal FAS~\cite{yu2023flexible} has attracted more attention, which aims to develop a unified multimodal FAS model using complete multimodal face data but is insensitive to test-time missing modalities. In this paper, we tackle one main challenge in flexible-modal FAS, i.e., when missing modality occurs either during training or testing in real-world situations. Inspired by the recent success of the prompt learning in language models, we propose \textbf{V}isual \textbf{P}rompt flexible-modal \textbf{FAS} (VP-FAS), which learns the modal-relevant prompts to adapt the frozen pre-trained foundation model to downstream flexible-modal FAS task. Specifically, both vanilla visual prompts and residual contextual prompts are plugged into multimodal transformers to handle general missing-modality cases, while only requiring less than 4\% learnable parameters compared to training the entire model. Furthermore, missing-modality regularization is proposed to force models to learn consistent multimodal feature embeddings when missing partial modalities. Extensive experiments conducted on two multimodal FAS benchmark datasets demonstrate the effectiveness of our VP-FAS framework that improves the performance under various missing-modality cases while alleviating the requirement of heavy model re-training. 
\end{abstract}


\IEEEpeerreviewmaketitle

\section{Introduction}

%\IEEEPARstart{G}{esture} recognition is attracting more and more attention in both research and industrial communities due to its vast applications~\cite{weissmann1999gesture,rautaray2015vision,liu20203d,liu20203dsk}. The main issue in gesture recognition is to learn spatio-temporal representations. To date, various 3D convolutional neural networks (3DCNN) have been proposed to address this issue~\cite{tran2015learning,qiu2017learning,carreira2017quo,xie2018rethinking}. However, existing variants are still hard to model both the local subtle motion and global dynamics, and most of them have to turn to optical flow~\cite{ilg2017flownet,sun2018pwc} for better modelling. The drawbacks of previous 3DCNN variants are obvious in the case of multi-modalities as the current hand-crafted architectures have low efficiency in joint learning~\cite{wang2018rgb}. In this paper, we will revisit  spatio-temproal feature learning problem in a multi-modal situation.  

\IEEEPARstart{F}{ace} recognition technology has been widely used in many intelligent systems due to its convenience and remarkable accuracy. However, face recognition systems are still vulnerable to presentation attacks (PAs) ranging from print, replay, and 3D-mask attacks. Therefore, both academia and industry have recognized the critical role of face anti-spoofing (FAS)~\cite{yu2021deep} in securing the face recognition system.

In the past decade, plenty of hand-crafted features based~\cite{boulkenafet2015face,Boulkenafet2017Face,Komulainen2014Context,Patel2016Secure} and deep learning based~\cite{qin2019learning,Liu2018Learning,yang2019face,Atoum2018Face,Gan20173D,george2019deep,li2021asymmetric,zhang2022unsupervised,qiao2022fgdnet} methods have been proposed for unimodal FAS. Despite satisfactory performance in seen attacks and environments, unimodal methods generalize poorly on emerging novel attacks and unseen deployment conditions. Thanks to the advanced imaging sensors with various modalities (e.g., RGB, Infrared (IR), Depth, Thermal)~\cite{george2019biometric}, multimodal methods facilitate the FAS applications under some high-security scenarios with low false acceptance errors (e.g., face payment and vault entrance guard).

\begin{figure}
\centering
%\vspace{-0.8em}
\includegraphics[scale=0.38]{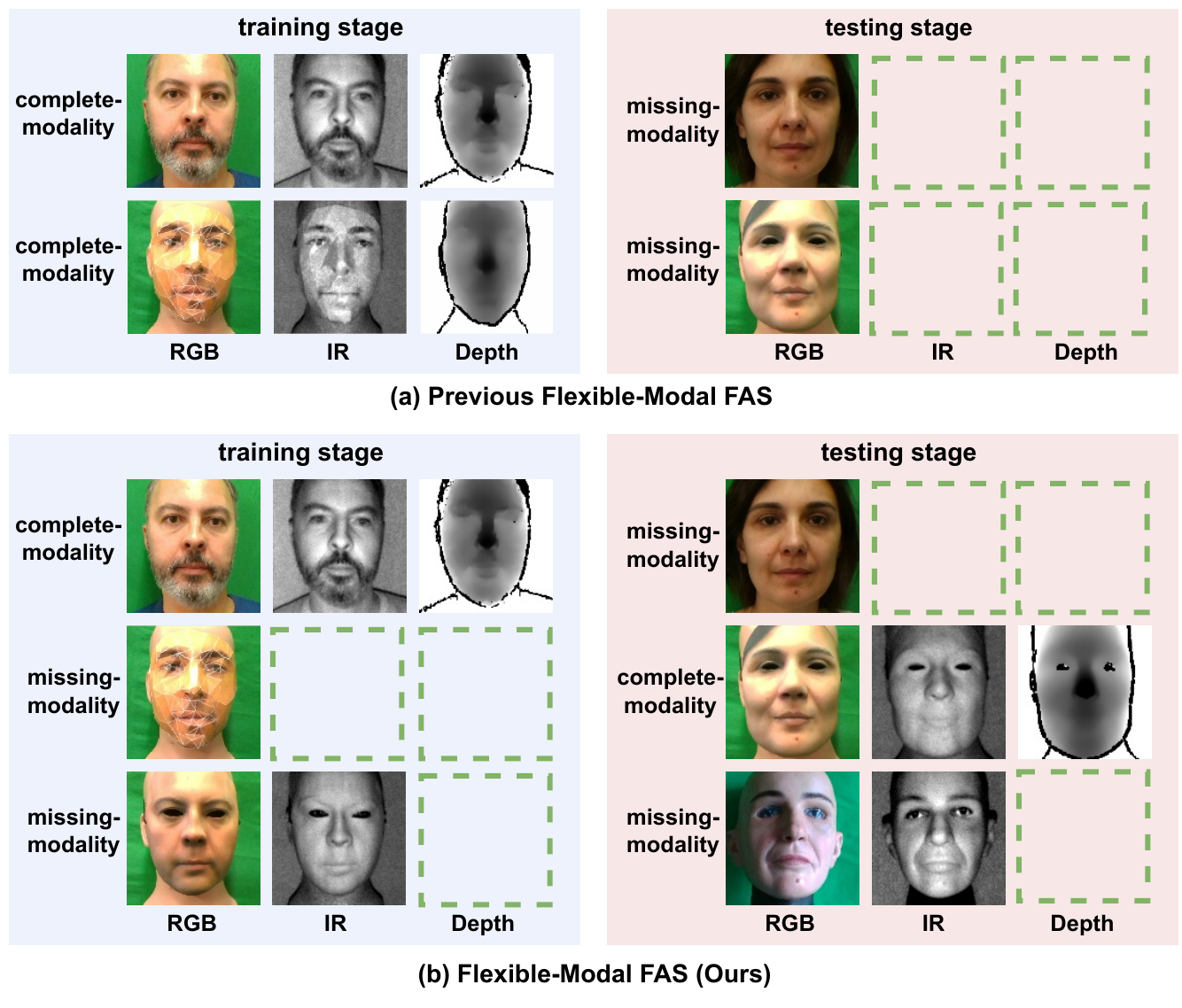}
\vspace{-0.8em}
  \caption{\small{
  Illustration of our proposed flexible-modal FAS settings. Prior work~\cite{yu2023flexible} investigates the robustness of multimodal FAS models to missing-modality test data using complete-modality training data. In contrast, our work studies a more general scenario where missing-modality cases occur differently not only for each data sample but also at arbitrary stages (training, testing, or both).
  }
}
\label{fig:Figure1}
%\vspace{-1.0em}
\end{figure}   

Recently, due to the strong long-range and cross-modal representation capacity, vision transformer (ViT)~\cite{dosovitskiy2020image} based methods~\cite{liuma,george2020effectiveness} have been proposed to improve the robustness of FAS systems. However, on the one hand, most multimodal transformer-based FAS methods~\cite{yu2023rethinking} assume that complete multimodal data are always available, which may not hold in practice due to the constraints of users' devices and imaging sensors. Thus, the performance may degrade sharply when the missing-modality~\cite{yu2023rethinking} scenarios occur during the training or testing phases.

On the other hand, due to the task and modality gaps~\cite{bachmann2022multimae} between RGB-based generic object classification and multimodal FAS, it is still challenging to efficiently finetune ImageNet pretrained transformers for downstream tasks~\cite{george2020effectiveness}. From the perspective of computation cost, the size and inference time of transformer-based foundation models increases thus finetuning becomes significantly expensive, which is not even feasible for practitioners due to the limited computation resources in most real-world applications. In addition, from the perspective of data scale, finetuning a transformer on a downstream task with relatively small-scale multimodal FAS datasets might result in overfitting and limited generalizability.

This motivates us to design a method that allows multimodal transformers to alleviate these two real-world challenges. Some pioneer works~\cite{yu2023flexible,liuma,george2021cross} investigate the sensitivity of well-trained multimodal FAS models against the presence of modal-incomplete test data (i.e., specific IR, Depth, or Depth-IR modalities are missing). However, they only consider the case of missing specific modalities for all the data samples in the inference stage, while in real-world scenarios 1) the missing modality for each input data could not be known in advance. In other words, partial complete-modality data with other missing-modality data would make more sense; and 2) due to the variance among collected imaging sensors, missing modality might occur in both training and testing stages.

In this paper, we study flexible-modal FAS under more general modality-incomplete and modality-complete scenarios, where various missing-modality cases may occur in any data samples during both training and testing stages. For example, there can be both RGB-only and RGB+Depth data during training and testing. In particular, we also focus on alleviating the requirement of finetuning the entire multimodal FAS model. To this end, we propose a framework with advanced prompt learning techniques for addressing the aforementioned challenges. Basically, prompt learning methods~\cite{jia2022visual,zhu2023visual,lee2023multimodal} emerge recently as efficient and effective solutions for adapting pre-trained transformers to the target domain via only training very few parameters in learnable prompts and achieve comparable performance with finetuning the whole heavy model. As motivated by~\cite{yu2020multi} which shows that multimodal central difference cues benefit robust FAS, besides vanilla prompts, we also design residual contextual prompts into multimodal transformers to handle general missing-modality cases. Specifically, multimodal central difference contexts and multi-level residual contexts are utilized for guiding residual contextual prompt learning, which is able to represent intrinsic FAS-related cues. The size of our learnable prompts can be less than 4\% of the entire transformer, and thus the computation becomes more affordable compared to holistic finetuning. The key differences between our work and~\cite{yu2023flexible,liuma,george2021cross} are illustrated in Figure~\ref{fig:Figure1}.

In order to further enhance the robustness of the multimodal FAS models under missing-modality scenarios, we propose the missing-modality regularization to guide the deep multimodal embedding features as similar as possible in cases of complete modalities and missing partial modalities. In this way, models' sensitivity to missing modalities can be alleviated while keeping rich complete-modality representation capacity. To sum up, our main contributions are as follows:
\begin{itemize}
\setlength\itemsep{-0.1em}
\vspace{-0.5em}
    \item We introduce novel scenarios for flexible-modal FAS, where the missing modality may occur differently for each data sample, either in \textit{training} or \textit{testing} phase. Compared with previous flexible-modal FAS works only focusing on test-time missing modality, the proposed flexible-modal FAS protocols are more general and toward real-world applications.

    \item  We propose \textbf{V}isual \textbf{P}rompt flexible-modal \textbf{FAS} (VP-FAS) to plug both vanilla visual prompts and residual contextual prompts into multimodal transformers to handle general missing-modality cases, while only requiring 3.6\% parameters to adapt pre-trained models, thus avoiding finetuning heavy transformers. 
    
    \item  We propose the missing-modality regularization to force models to learn consistent multimodal feature embeddings when missing partial modalities. 

    \item Our proposed methods achieve state-of-the-art performance with most of the missing-modality and complete-modality settings on both intra- as well as cross-dataset testings.

\end{itemize}

%\vspace{-1.3em}
\section{Related Work}
%\vspace{-0.3em}

%In this section, we first introduce recent progresses in unimodal and multimodal FAS. Then variants of vision transformer and recent parameter-efficient tuning methods will be reviewed.

\vspace{0.5em}
\noindent\textbf{Unimodal face anti-spoofing.}\quad      
Traditional FAS methods usually extract handcrafted features~\cite{boulkenafet2015face,Komulainen2014Context} from the facial images to capture the spoofing patterns. Recently, a few deep learning-based methods are proposed for FAS. On the one hand, FAS can be naturally treated as a binary classification task, thus binary cross-entropy loss is used for model supervision. On the other hand, according to the physical discrepancy between live and spoof faces, dense pixel-wise supervisions~\cite{yu2021revisiting} such as pseudo depth map~\cite{Liu2018Learning,yu2020searching,wang2020deep}, reflection map~\cite{yu2020face}, texture map~\cite{zhang2020face} and binary map~\cite{george2019deep} are designed for fine-grained learning. Moreover, to detect unknown attacks under unseen domains successfully, domain generalization~\cite{jia2020single,wang2022domain}  and zero-shot learning~\cite{liu2019deep,qin2019learning} methods are developed.

\begin{figure*}
\centering
%\vspace{-0.8em}
\includegraphics[scale=0.35]{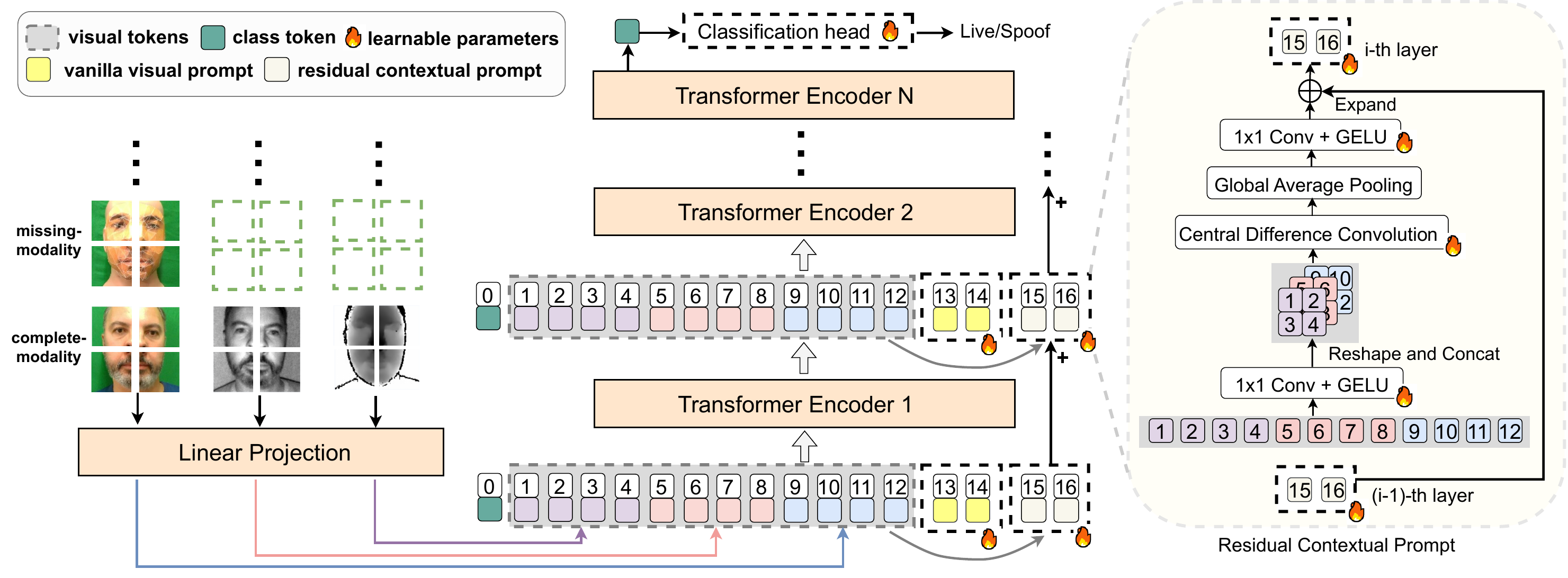}
\vspace{-0.3em}
  \caption{\small{
  The framework of the Visual Prompt flexible-modal FAS (VP-FAS). The vanilla visual prompts, residual contextual prompts, and classification head are \textbf{trainable} while the linear projection and vanilla transformer encoder blocks are fixed with the pre-trained parameters. To learn modality-aware task-related prompts, multimodal central difference contexts and adjacent-level residual contexts are utilized for guiding residual contextual prompt learning.
  }
}
\label{fig:VP-FAS}
\vspace{-0.3em}
\end{figure*}

\vspace{0.5em}
\noindent\textbf{Multimodal face anti-spoofing.}\quad      
With multimodal inputs (e.g., RGB, IR, Depth, and Thermal), there are a few multimodal FAS works that consider input-level~\cite{nikisins2019domain,liu2021data,george2020learning,nikisins2019domain} and decision-level~\cite{zhang2019feathernets} fusions. Besides, mainstream FAS methods extract complementary multi-modal features using feature-level fusion~\cite{yu2020multi,zhang2019dataset,liu2021face,wang2022conv,liuma,li2021asymmetric} strategies. As there is redundancy across multi-modal features, direct feature concatenation~\cite{yu2020multi} easily results in high-dimensional features and overfitting. To alleviate this issue, feature re-weighting mechanism~\cite{zhang2019dataset,zhang2020casia} and Modal Feature Erasing operation~\cite{shen2019facebagnet} are proposed to mine informative multimodal cues and alleviate modality-aware overfitting. Beside complete multimodal FAS, a few methods such as cross-attention~\cite{yu2023flexible}, cross-modal focal loss~\cite{george2021cross}, and modality-agnostic ViT~\cite{liuma,liu2023fm} are designed for flexible-modal FAS when trained with complete-modality data and tested on complete- or missing-modality data. Different from previous flexible-modal FAS only focusing on test-time missing modality, we introduce novel scenarios where the missing modality may occur differently for each data sample in the training and testing phases.

\vspace{0.5em}
\noindent\textbf{Transformer for vision tasks.}\quad  
Transformer~\cite{lin2021survey} is proposed in~\cite{vaswani2017attention} to model sequential data in the field of NLP. Then vision transformer (ViT)~\cite{dosovitskiy2020image} is proposed recently by feeding the transformer with sequences of image patches for image classification. Many other ViT variants ~\cite{han2020survey,khan2021transformers,touvron2021training,liu2021swin,wang2021pyramid,han2021transformer,chen2021crossvit,ding2021hr} are proposed from then, which achieve promising performance compared with its counterpart CNNs for image analysis tasks~\cite{carion2020end,zheng2021rethinking,he2021transreid}. Recently, some works introduce ViT for face attack detection tasks such as face anti-spoofing~\cite{liuma,ming2022vitranspad,george2020effectiveness,wang2022face,wang2022learning,yu2021transrppg} and face forgery detection~\cite{kong2023enhancing,miao2023f}. 
Meanwhile, multimodal transformers have become popular for vision-related multimodal tasks, such as audio-visual recognition~\cite{nagrani2021attention}, vision language understanding~\cite{zheng2022general}, Bird's-eye-view (BEV)~\cite{qin2022uniformer}, and multi-spectral vision~\cite{liuma,zhu2023visual,yu2023rethinking} tasks.

\vspace{0.5em}
\noindent\textbf{Parameter-efficient tuning.}\quad      Parameter-efficient tuning focuses on adapting the pre-trained model on a downstream task with a few parameters. Two lines of parameter-efficient tuning approaches have been proposed recently. On the one hand, learning task-relevant information by applying prompts~\cite{jia2022visual,hu2021lora,lee2023multimodal,yang2022prompting} to the input tokens or designing lightweight adapters~\cite{houlsby2019parameter,chen2022adaptformer} to adapt pre-trained information have acquired promising results for the performance and efficiency. On the other hand, distribution alignment~\cite{lian2022scaling} between pre-trained and downstream tasks has been shown to be efficient. Few works~\cite{zhu2023visual,sung2022vl} also consider parameter-efficient tuning on ImageNet pretrained ViT for multimodal tasks, aiming at efficiently learning modality-complementary cues. However, there are still no parameter-efficient tuning works for flexible-modal FAS.  

\vspace{-0.3em}
\section{Methodology}
\label{sec:method}

In this section, we will first define the problem setting of ViT-based flexible-modal FAS in Sec.~\ref{sec:ViT}, then introduce visual prompt and residual contextual prompt tuning framework in Sec.~\ref{sec:VP-FAS}, and at last present the missing-modality regularization in Sec.~\ref{sec:MMR}. 

\vspace{-0.3em}
\subsection{ViT-based Flexible-Modal FAS}
\label{sec:ViT}

Due to the powerful long-range relation modeling~\cite{george2020effectiveness,wang2022face} and multimodal reasoning~\cite{liuma} capacity of transformer models, here we focus on ViT-based flexible-modal FAS with real-world RGB and RGB+X (`X' indicates Depth, IR, thermal, etc.) imaging data.

To be a classical setting but without loss of generality, we consider the multimodal scenarios consisting of $M$=3 modalities, i.e., RGB modality, IR modality, and Depth (D) modality. Please note that other scenarios (e.g., $M$=2, RGB-IR, and RGB-D) are omitted here. Given a multimodal dataset $S=\left\{ S^{RGB-D-IR},S^{RGB-D},S^{RGB-IR},S^{RGB} \right\}$, we denote $ S^{RGB-D-IR}=\left\{ x^{RGB}_{i},x^{D}_{i},x^{IR}_{i},y_{i} \right\}$ as the modality-complete subset, while $ S^{RGB-D}=\left\{ x^{RGB}_{j},x^{D}_{j},y_{j} \right\}$ and $ S^{RGB-IR}=\left\{ x^{RGB}_{k},x^{IR}_{k},y_{k} \right\}$ are denoted as the modality-incomplete subsets where one modality is missing. Similarly, $S^{RGB}=\left\{ x^{RGB}_{l},y_{l} \right\}$ is denoted as the modality-incomplete RGB-only subset where two modalities are missing. As shown in Figure~\ref{fig:VP-FAS}, the training data may contain data samples with different missing cases including complete data $S^{RGB-D-IR}$ and modality-incomplete data (e.g., $S^{RGB}$).

For simplicity, we adopt the ImageNet-22K pretrained ViT-B~\cite{dosovitskiy2020image} as our backbone model, which consists of a patch tokenizer $\mathbf{E}^{\text{patch}}$ via linear projection, $N$=12 transformer layers $\mathbf{E}_{i}^{\text{Trans}}$ ($ i=1,..., N$) and a classification head $\mathbf{E}^{\text{Head}}$. The flexible-modal inputs (missing modalities are set as zeros) from $S$ are passed over $\mathbf{E}^{\text{patch}}$ to generate the visual tokens $T^\text{Vis}$, which is concatenated with a learnable class token $T^\text{Cls}$, and added with position embeddings. Then all patch tokens $T^\text{All}=[T^\text{Vis},T^\text{Cls}]$ will be forwarded with $\mathbf{E}^{\text{Trans}}$. Finally, $T^{\text{Cls}}$ is sent to $\mathbf{E}^{\text{Head}}$ for binary live/spoof prediction $\widehat{y}$.

\vspace{-1.3em}
\subsection{Visual Prompt Flexible-Modal FAS}
\label{sec:VP-FAS}

\noindent\textbf{Vanilla visual prompt learning.}\quad  Recent studies have verified that introducing visual prompts~\cite{jia2022visual} in shallow or all transformer layers can efficiently transfer ImageNet-22K pretrained knowledge to downstream tasks. Compared with direct finetuning entire backbone, visual prompt tuning has far fewer trainable parameters and converges more stably. Learning visual prompts for flexible-modal FAS has two advantages: 1) efficiently mining modality-aware semantics for flexible-modal scenarios; and 2) alleviating overfitting due to small-scale multimodal FAS data. 

Given a pretrained $N$-layer ViT model, we fix all the pre-trained parameters from $\mathbf{E}^{\text{Patch}}$ and $\mathbf{E}^{\text{Trans}}$ while training only visual prompts and $\mathbf{E}^{\text{Head}}$. Specifically, we introduce a set of $p$ continuous embeddings of dimension $d$, i.e., prompts, in the input space of every transformer layer. For $i$-th Layer $L_{i}$, we denote the collection of input learnable prompts (see yellow 13-th and 14-th tokens in Figure~\ref{fig:VP-FAS} for illustration) as $T^{\mathbf{P}}_{i}=\left\{ \mathbf{p}^{k}_{i}\in \mathbb{R}^{d} | k\in \mathbb{N},1\leqslant k\leqslant N \right\}$. The deep-prompted tokens for each layer are formulated as  
\begin{equation}
\begin{split}
\left [ T^\text{Cls}_{i},T^\text{Vis}_{i}, \setminus  \right ]&=\mathbf{E}_{i}^{\text{Trans}}\left ( \left [ T^\text{Cls}_{i-1},T^\text{Vis}_{i-1},T^{\mathbf{P}}_{i-1} \right ] \right ), \quad i=1,2,...,N, \\
\widehat{y}&=\mathbf{E}^{\text{Head}}\left ( T^\text{Cls}_{N} \right ).
\label{eq:prompt1}
\end{split}
\end{equation}

%\vspace{0.2em}
\noindent\textbf{Residual contextual prompt learning.}\quad   
Despite efficiency, vanilla learnable prompts still have three shortcomings for flexible-modal FAS: 1) limited contextual modality-aware interaction~\cite{yu2020multi,george2019biometric} for robust flexible-modal representation; 2) lack of local detailed spatial contexts~\cite{yu2020searching} for intrinsic spoofing cue mining; and 3) lack of multi-level context aggregation~\cite{Liu2018Learning}, which has been verified to be important in FAS task. As motivated by~\cite{yu2020multi} which shows that multimodal central difference cues benefit robust FAS, besides vanilla prompts, we also design $p$ residual contextual prompts into each transformer layer to handle general flexible-modality cases. Specifically, multimodal central difference contexts and multi-level residual contexts are utilized for guiding residual contextual prompt learning, which is able to represent intrinsic FAS-related cues.

As illustrated in Figure~\ref{fig:VP-FAS}, the residual contextual prompt $T^{\mathbf{P}\_RC}_{i}$ for the $i$-th transformer layer consists of three parts: 1) visual base prompt $T^{\mathbf{P}\_base}_{i}$; 2) multimodal central difference contexts $T^{\mathbf{P}\_cd}_{i}$; and 3) adjacent-level residual contexts $T^{\mathbf{P}\_RC}_{i-1}$ (only when $i\geq$2). Specifically, for the multimodal central difference contexts, a 1×1 convolution with GELU $\Theta_{\downarrow}$ is conducted on visual tokens $T^\text{Vis}$ for dimension reduction from the original channels $d$ to a hidden squeezed dimension $d'$, which is then spatially reshaped and modality-aware concatenated. Then a 3×3 central difference convolution~\cite{yu2020searching} $\Theta_{\text{CDC}}$ mapping channels $d'$×$M$ to $d'$ for multimodal local context representation, which cascades a global average pooling (GAP) layer for spatial context aggregation. Finally, an 1×1 convolution with GELU $\Theta_{\uparrow}$ for dimension expansion to $d$. To sum up, the multimodal central difference contexts can be formulated as
\begin{equation}
T^{\mathbf{P}\_cd}=\Theta_{\uparrow}(\text{GAP}(\Theta_{\text{CDC}}(\text{Concat}[\Theta_{\downarrow}(T^\text{Vis}_{\text{RGB}}),\Theta_{\downarrow}(T^\text{Vis}_{\text{IR}}),\Theta_{\downarrow}(T^\text{Vis}_{\text{D}})]))).
\label{eq:AMA}
\end{equation}
Therefore, the residual contextual prompt for each layer can be formulated as
\begin{equation}
\left\{\begin{matrix}
T^{\mathbf{P}\_RC}_{i}=T^{\mathbf{P}\_base}_{i}+\text{Expand}(T^{\mathbf{P}\_cd}_{i}), \quad i=1, \\
T^{\mathbf{P}\_RC}_{i}=T^{\mathbf{P}\_base}_{i}+\text{Expand}(T^{\mathbf{P}\_cd}_{i})+T^{\mathbf{P}\_RC}_{i-1},\quad 2\leq i\leq N,
\end{matrix}\right.
\end{equation}
where $\text{Expand}(\cdot)$ expands the dimension of multimodal central difference contexts up to the prompt length $p$.

\vspace{0.3em}
\noindent\textbf{Combing vanilla and residual contextual prompts.}\quad   
To leverage the advantages of global semantics and contextual relationship, we consider $p//$2 vanilla visual prompts and $p//$2 residual contextual prompts for each transformer layer. Overall, tokens for $i$-th layer in our proposed VP-FAS with both vanilla visual prompts and residual contextual prompts are formulated as
\begin{equation}
\left [ T^\text{Cls}_{i},T^\text{Vis}_{i}, \setminus , \setminus  \right ]=\mathbf{E}_{i}^{\text{Trans}}\left ( \left [ T^\text{Cls}_{i-1},T^\text{Vis}_{i-1},T^{\mathbf{P}}_{i-1}, T^{\mathbf{P}\_RC}_{i-1} \right ] \right ).
\label{eq:prompt1}
\end{equation}

\begin{figure}[t]
\centering
\includegraphics[scale=0.36]{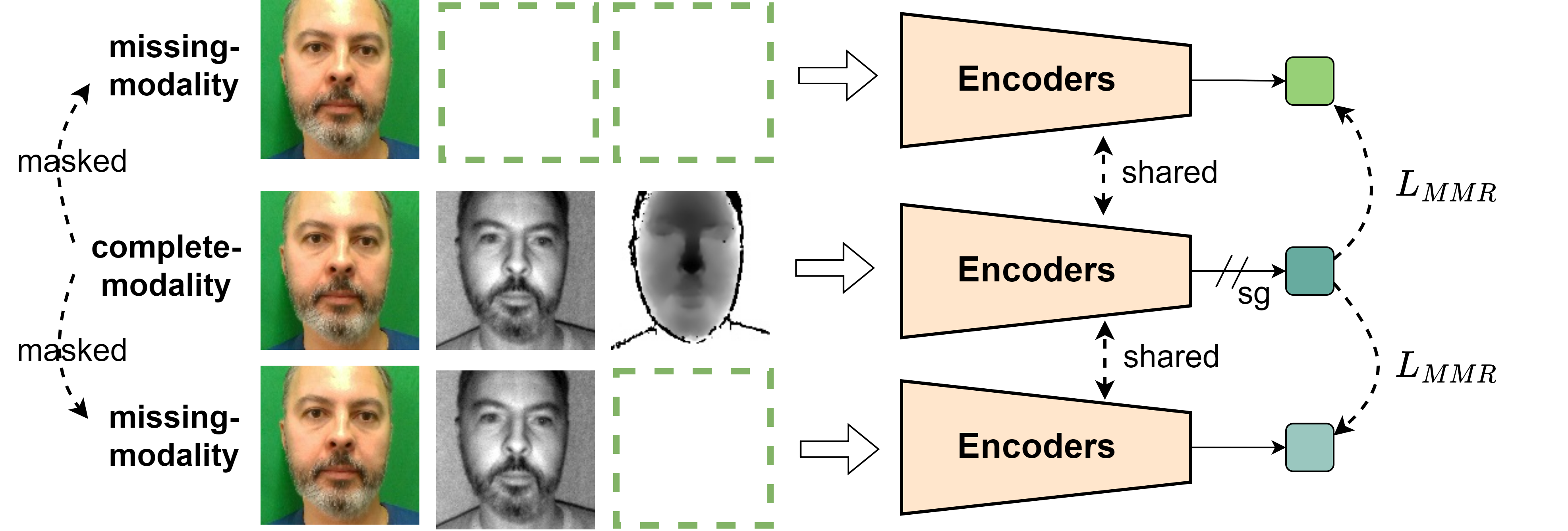}
\vspace{-0.3em}
  \caption{\small{Framework of the missing-modality regularization (MMR). Given multimodal inputs, missing-modality data can be generated via partial modality masking. Then both complete- and missing-modality data are passed over the shared encoders to obtain the embeddings (e.g., class token) as well as their MMR losses. `sg' is short for `stop gradient'.}
  }
\label{fig:MMR}
\vspace{-0.3em}
\end{figure}

\subsection{Missing-Modality Regularization}
\label{sec:MMR}
In order to further enhance the robustness of the multimodal FAS models under missing-modality scenarios, we propose the missing-modality regularization (MMR) to guide multimodal model learning, forcing the deep multimodal embedding features as similar as possible in cases of complete modalities and missing partial modalities. In this way, models' sensitivity to missing modalities can be alleviated while keeping rich complete-modality representation capacity.

As shown in Figure~\ref{fig:MMR}, given a modality-complete subset (e.g., $S^{RGB-D-IR}$ for the 3-modality case), missing-modality data can be generated via partial-modality masking (e.g., $\mathcal{M}_{D}$, $\mathcal{M}_{IR}$, and $\mathcal{M}_{D-IR}$). Here we set a hyperparameter $\gamma$ to control the masking ratio for each possible partial-modality missing situation except missing RGB. For example, when the masking of $\mathcal{M}_{D}$ occurs, both complete- and generated missing-depth data are passed over the shared transformer encoders to obtain their class token embeddings $T^\text{Cls}_{RGB-D-IR}$ and $T^\text{Cls}_{RGB-IR}$, respectively. Then negative embedding similarity is utilized as MMR loss, which can be formulated as
\begin{equation}
\mathcal{L}_{\text{MMR}}=-\frac{(T^\text{Cls}_{RGB-D})'}{\left\| T^\text{Cls}_{RGB-D}\right\|_{2}}\cdot sg(\frac{T^\text{Cls}_{RGB-D-IR}}{\left\| T^\text{Cls}_{RGB-D-IR}\right\|_{2}}),
\label{eq:prompt1}
\end{equation}
where $'$ and $sg(\cdot)$ indicates the transpose and stop gradient~\cite{grill2020bootstrap} operation, respectively. Finally, the overall loss for supervision is the summation of binary cross-entropy loss and MMR loss. Please note that we also try our MMR on the predicted logits using Kullback–Leibler divergence (KL) metric. However, we find that it is unstable and hard to converge.

%\vspace{-0.3em}
\section{Experimental Evaluation}
%\vspace{-0.2em}
\label{sec:experiemnts}

\subsection{Datasets and Performance Metrics}
\label{sec:dataset} 
%\vspace{-0.2em}

Two commonly used multimodal FAS datasets are used for experiments, including WMCA~\cite{george2019biometric} and CASIA-SURF~\cite{zhang2019dataset}. \textbf{WMCA} contains a wide variety of 2D and 3D PAs with four modalities, which introduces 2 protocols: the `seen' protocol which emulates the seen attack scenario, and the `unseen' attack protocol which evaluates the generalization on an unseen attack. \textbf{CASIA-SURF} consists of 1000 subjects with 21000 videos, and each sample has 3 modalities, which has an official intra-testing protocol. In this paper, we conduct intra- and cross-dataset flexible-modal testings on WMCA (`seen' protocol) and CASIA-SURF datasets, and leave flexible-modal FAS on `unseen' protocol of WMCA for future works.  

\begin{figure}[t]
\centering
\includegraphics[scale=0.5]{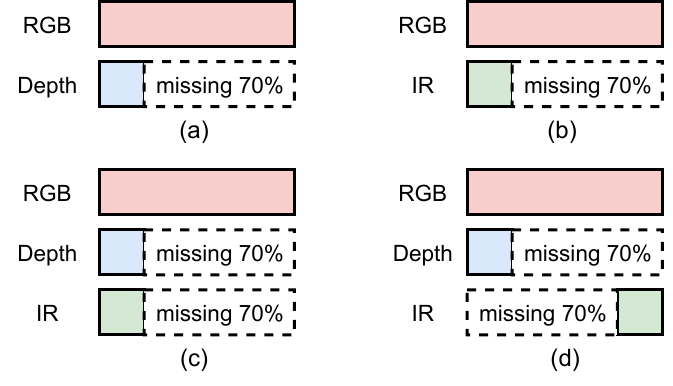}
\vspace{-0.3em}
  \caption{\small{Setting of flexible-modal FAS when missing-modality ratio $\alpha$=70\%. (a) RGB-D modalities when 70\% Depth modality is missing. (b) RGB-IR modalities when 70\% IR modality is missing. (c) RGB-D-IR modalities when overlapped 70\% Depth \& IR modalities are missing. (d) RGB-D-IR modalities when 70\% Depth \& IR modalities are missing but with limited complete RGB-D-IR data.}
  }
\label{fig:setting}
\vspace{-0.3em}
\end{figure} 

In terms of evaluation metrics, Attack Presentation Classification Error Rate (APCER), Bonafide Presentation Classification Error Rate (BPCER), and ACER~\cite{ACER} are used for the intra-dataset tests. The ACER on the testing set is determined by the Equal Error Rate (EER) threshold on dev sets for CASIA-SURF, and the BPCER=1\% threshold for WMCA. For cross-testing experiments, Half Total Error Rate (HTER) is adopted. 

\vspace{-0.3em}
\subsection{Setting of Flexible-Modal FAS}
\label{sec:setting} 
%\vspace{-0.2em}
Different from~\cite{yu2023flexible}, in this paper, we focus on the more flexible scenario where missing modalities might occur in both training and testing phases, where each modality for each data sample has chances to be lost. We define the missing-modality ratio $\alpha$ as the proportion of modality-incomplete data to the entire dataset. For the FAS task, we consider four practical settings (see Figure~\ref{fig:setting} for an example of $\alpha$=70\%) of missing modality: 1) RGB-D when missing Depth; 2) RGB-IR when missing IR; 3) RGB-D-IR when missing overlapped Depth and IR; RGB-IR when missing IR; and 4) RGB-D-IR when missing Depth and IR with limited complete RGB-D-IR data or limited overlapped D-IR data. For the first two cases, the missing-modality ratio $\alpha$ indicates that there are $\alpha$ RGB-only data and (1-$\alpha$) complete RGB-D and RGB-IR data, respectively. For the third case, there are $\alpha$ RGB-only data, and (1-$\alpha$) complete RGB-D-IR data. For the last case, when $\alpha <$50\%, there are $\alpha$ RGB-D and RGB-IR data, and (2*$\alpha$-100\%) complete RGB-D-IR data. In contrast, when 50\%$\leq\alpha<$100\%, there are (2*$\alpha$-100\%) RGB-only data, and (1−$\alpha$)\% RGB-D and RGB-IR data.

All training, validation, and testing sets follow the same flexible settings for each experiment.
In our main experiments in Tables~\ref{tab:intra1},~\ref{tab:intra2},~\ref{tab:cross1}, and \ref{tab:cross2}, the missing-modality ratio $\alpha$ is set to 70\% by default. We also illustrate the quantitative results of different $\alpha$ in Figures~\ref{fig:intra1},~\ref{fig:intra2},~\ref{fig:cross1} and~\ref{fig:cross2}.

%\vspace{-0.5em}
\subsection{Implementation Details}
%\vspace{-0.2em}
\label{sec:Details}

MTCNN~\cite{zhang2016joint} face detector is used for cropping face frames. Similar to~\cite{yu2023rethinking}, composition inputs `GRAY\_HOG\_PLGF' are adopted on unimodal and multimodal experiments for IR modality, while the raw inputs are utilized for RGB and Depth modalities. ViT-Base~\cite{dosovitskiy2020image} supervised by binary cross-entropy loss is used as the defaulted architecture. For direct finetuning, only the last transformer block and classification head are trainable. For our proposed VP-FAS, total prompt length $p$=40 is adopted. In other words, 20 vanilla prompts, as well as 20 residual contextual prompts, are used. Central difference intensity $\theta$=0.5 is the defaulted setting for $\Theta_{\text{CDC}}$ in residual contextual prompts. The original and hidden channels are $d$=768 and $d'$=64, respectively. For missing-modality regularization, the mask ratio $\gamma$=0.15 is used. The experiments are implemented with Pytorch on one NVIDIA V100 GPU. Adam optimizer with the fixed lr=2e-4, wd=5e-3, and batch size 16 is used at the training stage. We finetune models with a maximum of 40 epochs based on the ImageNet-22K pretrained ViT-B/16 weights.

\begin{table}[t]
\centering
\caption{ Intra-dataset results with $\alpha$=70\% on WMCA. All performance is evaluated using ACER(\%)$\downarrow$. Best results are marked in \textbf{bold}.} 
\vspace{-0.2em}\label{tab:intra1}
\resizebox{0.49\textwidth}{!}{\begin{tabular}{|c|c|c|c|c|}
\hline
Method                 & \begin{tabular}[c]{@{}c@{}}RGB-D\\ (missing 70\% D)\end{tabular} & \begin{tabular}[c]{@{}c@{}}RGB-IR\\ (missing 70\% IR)\end{tabular} & \begin{tabular}[c]{@{}c@{}}RGB-D-IR\\ (missing overlapped\\ 70\% D\&IR)\end{tabular} & \begin{tabular}[c]{@{}c@{}}RGB-D-IR\\ (missing 70\% D\&IR \\ with limited overlapping)\end{tabular} \\ \hline
ViT~\cite{dosovitskiy2020image}                   & 10.61                                                                & 5.86                                                               & 8.70                                                                                    & 6.37                                                                                                    \\ \hline
Prompt~\cite{jia2022visual}                 & 8.70                                                                  & 5.52                                                               & 5.84                                                                                   & 7.21                                                                                                    \\ \hline
CMFL~\cite{george2021cross}                   & 6.53                                                                 & 4.95                                                               & 5.08                                                                                   & 5.37                                                                                                    \\ \hline
CrossAtten~\cite{yu2023flexible}             & 8.47                                                                 & 5.29                                                               & 8.33                                                                                   & 7.15                                                                                                    \\ \hline
\textbf{VP-FAS (Ours)} & \textbf{3.67}                                                        & \textbf{3.91}                                                      & \textbf{4.54}                                                                          & \textbf{4.83}                                                                                           \\ \hline
\end{tabular}}
\vspace{-0.3em}
\end{table}

\begin{table}[t]
\centering
\caption{ Intra-dataset results with $\alpha$=70\% on CASIA-SURF. All performance is evaluated using ACER(\%)$\downarrow$. } 
\vspace{-0.2em}\label{tab:intra2}
\resizebox{0.49\textwidth}{!}{\begin{tabular}{|c|c|c|c|c|}
\hline
Method                 & \begin{tabular}[c]{@{}c@{}}RGB-D\\ (missing 70\% D)\end{tabular} & \begin{tabular}[c]{@{}c@{}}RGB-IR\\ (missing 70\% IR)\end{tabular} & \begin{tabular}[c]{@{}c@{}}RGB-D-IR\\ (missing overlapped\\ 70\% D\&IR)\end{tabular} & \begin{tabular}[c]{@{}c@{}}RGB-D-IR\\ (missing 70\% D\&IR \\ with limited overlapping)\end{tabular} \\ \hline
ViT~\cite{dosovitskiy2020image}                    & 20.78                                                            & \textbf{18.55}                                                     & 16.61                                                                                & 18.14                                                                                               \\ \hline
Prompt~\cite{jia2022visual}                 & 16.90                                                             & 22.04                                                              & 18.05                                                                                & 19.71                                                                                               \\ \hline
CMFL~\cite{george2021cross}                   & 17.73                                                            & 21.67                                                              & 17.35                                                                                & 19.42                                                                                               \\ \hline
CrossAtten~\cite{yu2023flexible}              & 22.43                                                            & 19.15                                                              & 16.94                                                                                & 16.82                                                                                               \\ \hline
\textbf{VP-FAS (Ours)} & \textbf{12.41}                                                   & 19.03                                                              & \textbf{14.63}                                                                       & \textbf{16.61}                                                                                      \\ \hline

\end{tabular}}
\vspace{-0.3em}
\end{table}

\vspace{-0.5em}
\subsection{Main Results}
In this subsection, we show results of $\alpha$=70\% on WMCA dataset with 1) four flexible-modal settings, and 2) our proposed VP-FAS and four baseline methods including vanilla multimodal ViT (ViT)~\cite{dosovitskiy2020image}, visual prompt tuning (Prompt)~\cite{jia2022visual}, cross-modal focal loss (CMFL)~\cite{george2021cross}, and cross attention (CrossAtten)~\cite{yu2023flexible}. 

\vspace{0.3em}
\noindent\textbf{Intra testing on WMCA~\cite{george2019biometric}.} \quad   
As shown in Table~\ref{tab:intra1}, we find that the proposed VP-FAS consistently improves four baselines by a convincing margin (at least 0.5\% ACER decrease) in all the scenarios, indicating that our combined vanilla and residual contextual prompts as well as the missing-modality regularization, without entire model finetuning, are able to tackle general missing-modality cases and provide rich modality-aware live/spoof cues. Besides, we also find from the results `Prompt' that the vanilla prompt tuning cannot stably outperform the ViT baseline in all settings due to the limited flexible-modal contextual modeling capacity.

\vspace{0.3em}
\noindent\textbf{Intra testing on CASIA-SURF~\cite{zhang2019dataset}.} \quad   
Similar to the results in WMCA, it can be seen from Table~\ref{tab:intra2} that the proposed VP-FAS outperforms `Prompt', `CMFL', and `CrossAtten' by large margins on the CASIA-SURF dataset under all flexible-modal settings, showing the effectiveness of modality-aware contextual prompt learning. However, we still find the issues of advanced multimodal learning strategies (i.e., `Prompt', `CMFL', `CrossAtten', and `VP-FAS') from the results in the third column. Obviously, vanilla ViT even performs better than all other methods in terms of flexible-modal RGB-IR setting. The reason might be caused by the noisy IR imaging data in CASIA-SURF influencing the robust flexible-modal feature representation learning.   

\vspace{0.3em}
\noindent\textbf{Cross testing from WMCA to CASIA-SURF.} \quad   
Besides evaluating models' discrimination ability under flexible-modal intra-dataset testings, it is necessary to evaluate their generalization capacity under domain shifts. The cross-dataset results when trained on WMCA and tested on CASIA-SURF are shown in Table~\ref{tab:cross1}. It is clear that 1) compared with baseline `ViT' and `Prompt', introducing visual prompts can significantly reduce HTER (at least 2\%) under all flexible settings; and 2) based on the foundation of visual prompt tuning, `VP-FAS' benefits from extra spatial-aware and modality-aware live/spoof cues modeling of residual contextual prompts, and further improves performance (at least 3\% HTER decrease). 

\vspace{0.3em}
\noindent\textbf{Cross testing from CASIA-SURF to WMCA.} \quad   
As shown in Table~\ref{tab:cross2}, compared with cross testing from WMCA to CASIA-SURF, multimodal FAS models trained on CASIA-SURF and then tested on WMCA usually perform worse. The reasons are in two folds: 1) the visual quality of face images in CASIA-SURF is lower than those in WMCA, and 2) the diversity of spoof attacks in CASIA-SURF is fewer than those in WMCA. Thus, multimodal models trained on lower-quality fewer-diversity are challenging when tested on unseen higher-quality, and complex novel attacks. We still can find that `VP-FAS' outperform baselines `ViT' and `Prompt' by huge margins especially under the RGB-D-IR setting with missing overlapped 70\% D\&R, indicating the good generalization capacity of the proposed flexible-modal method.

\begin{table}[t]
\centering
\caption{ Cross-dataset results with $\alpha$=70\% when trained on WMCA and tested on CASIA-SURF. All performance is evaluated using HTER(\%)$\downarrow$. Best results are marked in \textbf{bold}.} 
\vspace{-0.2em}\label{tab:cross1}
\resizebox{0.49\textwidth}{!}{\begin{tabular}{|c|c|c|c|c|}
\hline
Method                 & \begin{tabular}[c]{@{}c@{}}RGB-D\\ (missing 70\% D)\end{tabular} & \begin{tabular}[c]{@{}c@{}}RGB-IR\\ (missing 70\% IR)\end{tabular} & \begin{tabular}[c]{@{}c@{}}RGB-D-IR\\ (missing overlapped\\ 70\% D\&IR)\end{tabular} & \begin{tabular}[c]{@{}c@{}}RGB-D-IR\\ (missing 70\% D\&IR \\ with limited overlapping)\end{tabular} \\ \hline
ViT~\cite{dosovitskiy2020image}                    & 35.88                                                            & 40.4                                                               & 30.36                                                                                & 32.68                                                                                               \\ \hline
Prompt~\cite{jia2022visual}                 & 31.42                                                            & 34.44                                                              & 26.19                                                                                & 30.08                                                                                               \\ \hline
CMFL~\cite{george2021cross}                   & 28.68                                                            & 36.58                                                              & 29.59                                                                                & 32.03                                                                                               \\ \hline
CrossAtten~\cite{yu2023flexible}              & 27.08                                                            & 37.83                                                              & 27.71                                                                                & 30.23                                                                                               \\ \hline
\textbf{VP-FAS (Ours)} & \textbf{23.76}                                                   & \textbf{30.92}                                                     & \textbf{21.21}                                                                       & \textbf{26.59}                                                                                      \\ \hline
\end{tabular}}
\vspace{-0.3em}
\end{table}

\begin{table}[t]
\centering
\caption{ Cross-dataset results with $\alpha$=70\% when trained on CASIA-SURF and tested on WMCA. All performance is evaluated using HTER(\%)$\downarrow$. Best results are marked in \textbf{bold}.} 
\vspace{-0.2em}\label{tab:cross2}
\resizebox{0.49\textwidth}{!}{\begin{tabular}{|c|c|c|c|c|}
\hline
Method                 & \begin{tabular}[c]{@{}c@{}}RGB-D\\ (missing 70\% D)\end{tabular} & \begin{tabular}[c]{@{}c@{}}RGB-IR\\ (missing 70\% IR)\end{tabular} & \begin{tabular}[c]{@{}c@{}}RGB-D-IR\\ (missing overlapped\\ 70\% D\&IR)\end{tabular} & \begin{tabular}[c]{@{}c@{}}RGB-D-IR\\ (missing 70\% D\&IR \\ with limited overlapping)\end{tabular} \\ \hline
ViT~\cite{dosovitskiy2020image}                    & 42.83                                                            & 37.97                                                              & 36.4                                                                                 & 38.21                                                                                               \\ \hline
Prompt~\cite{jia2022visual}                 & 38.98                                                            & 34.07                                                              & 35.84                                                                                & 35.66                                                                                               \\ \hline
CMFL~\cite{george2021cross}                   & 39.23                                                            & 35.23                                                              & 33.32                                                                                & 36.64                                                                                               \\ \hline
CrossAtten~\cite{yu2023flexible}              & 40.64                                                            & 36.99                                                              & 34.74                                                                                & 37.03                                                                                               \\ \hline
\textbf{VP-FAS (Ours)} & \textbf{36.93}                                                   & \textbf{32.24}                                                     & \textbf{29.6}                                                                        & \textbf{34.84}                                                                                      \\ \hline

\end{tabular}}
\vspace{-0.1em}
\end{table}

%\vspace{-0.3em}
\subsection{Ablation Study}
\label{sec:ablation}
We provide the ablation results for important components of residual contextual prompt and missing-modality regularization on WMCA with flexbile-modal RGB-D and RGB-IR settings.

\vspace{0.3em}
\noindent\textbf{Impact of the residual contextual prompt.}\quad   As shown in the first five rows of Table~\ref{tab:ablation}, compared with `ViT', `ViT+Vanilla Prompt' decreases 1.91\% ACER on flexible-modal RGB-D setting when missing 70\% Depth data, indicating the modality-aware modeling ability from learnable prompts. In contrast, the proposed `ViT+Contextual Prompt' performs slightly better than `ViT+Vanilla Prompt' on both flexible-modal RGB-D and RGB-IR settings. The reason is that contextual prompts leverage multimodal central difference contexts $T^{\mathbf{P}\_cd}$ and are beneficial for intrinsic live/spoof pattern representation. Furthermore, We can find from `ViT+Residual Contextual Prompt' that aggregating adjacent-level contextual prompts can make model learning more stable, and improve flexible-modal performance on both RGB-D and RGB-IR settings. Finally, it can be seen from `VP-FAS w/o MMR' that learning with both vanilla and residual contextual prompts boosts flexible-modal performance significantly, indicating the excellent global and contextual complementarity between these two kinds of prompts.

\begin{table}[t]
\centering
\caption{ Ablation results (ACER(\%)$\downarrow$) on WMCA with $\alpha$=70\%.} 
\vspace{-0.2em}\label{tab:ablation}
\resizebox{0.42\textwidth}{!}{\begin{tabular}{|c|c|c|c|c|}
\hline
Method                         & \begin{tabular}[c]{@{}c@{}}RGB-D\\ (missing 70\% D)\end{tabular} & \begin{tabular}[c]{@{}c@{}}RGB-IR\\ (missing 70\% IR)\end{tabular} \\ \hline
ViT                            & 10.61                                                            & 5.86                                                               \\ \hline
ViT+Vanilla Prompt             & 8.70                                                              & 5.52                                                               \\ \hline
ViT+Contextual Prompt          & 8.05                                                             & 5.09                                                               \\ \hline
ViT+Residual Contextual Prompt & 7.43                                                             & 4.83                                                               \\ \hline
VP-FAS w/o MMR                 & 5.96                                                             & 4.36                                                               \\ \hline
VP-FAS w/o `sg' in MMR                 & 10.47                                                            & 6.78                                                               \\ \hline
\textbf{VP-FAS (Ours)}         & \textbf{3.67}                                                    & \textbf{3.91}                                                      \\ \hline

\end{tabular}}
\vspace{-0.8em}
\end{table}

 \begin{figure}[t]
\centering
%\vspace{-1.0em}
\includegraphics[scale=0.41]{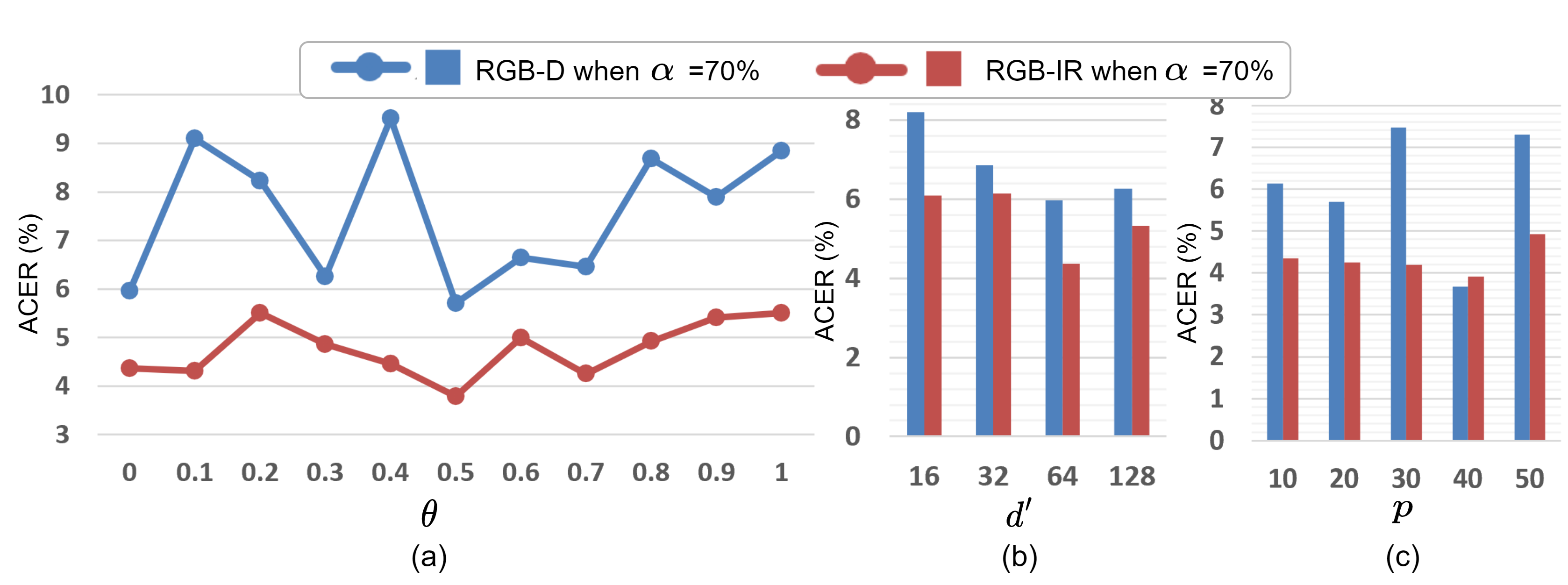}
\vspace{-0.8em}
  \caption{\small{
 Ablations on WMCA of (a) central difference intensity $\theta$ and (b) hidden dimension $d'$ in the residual contextual prompt; and (c) prompt length $p$ in VP-FAS. }
  }
\label{fig:ablation}
\vspace{-0.6em}
\end{figure}

\begin{figure*}[t]
\centering
%\vspace{-0.5em}
\includegraphics[scale=0.41]{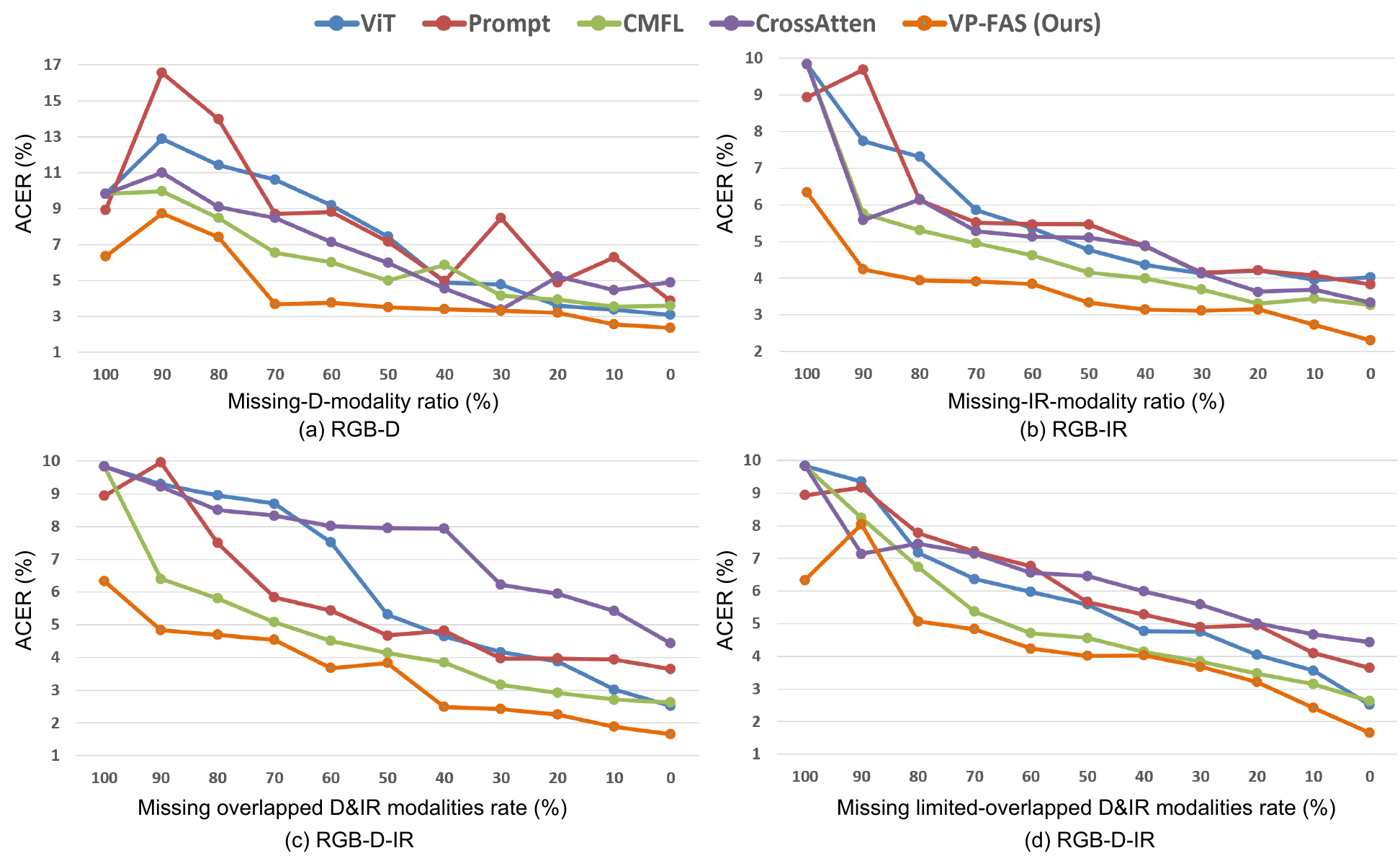}
\vspace{-0.8em}
  \caption{\small{
 Quantitative intra-testing results on WMCA with different modality-missing ratios under four flexible-modal settings. }
  }
\label{fig:intra1}
\vspace{-0.3em}
\end{figure*}

\begin{figure*}
\centering
%\vspace{-0.5em}
\includegraphics[scale=0.41]{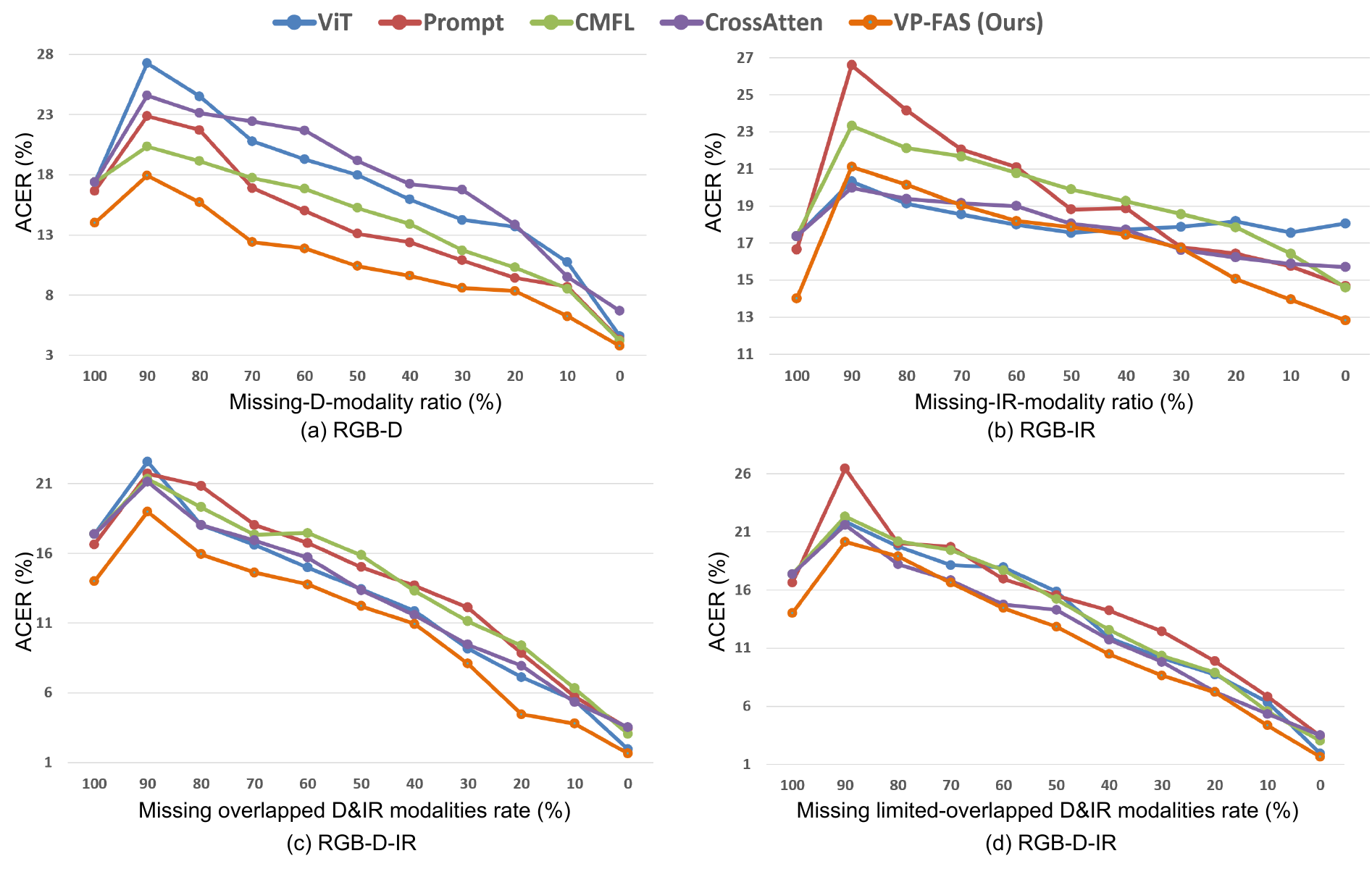}
\vspace{-0.8em}
  \caption{\small{
 Quantitative intra-testing results on CASIA-SURF with different modality-missing ratios under four flexible-modal settings. }
  }
\label{fig:intra2}
\vspace{-0.3em}
\end{figure*}

\begin{figure*}
\centering
%\vspace{-1.5em}
\includegraphics[scale=0.41]{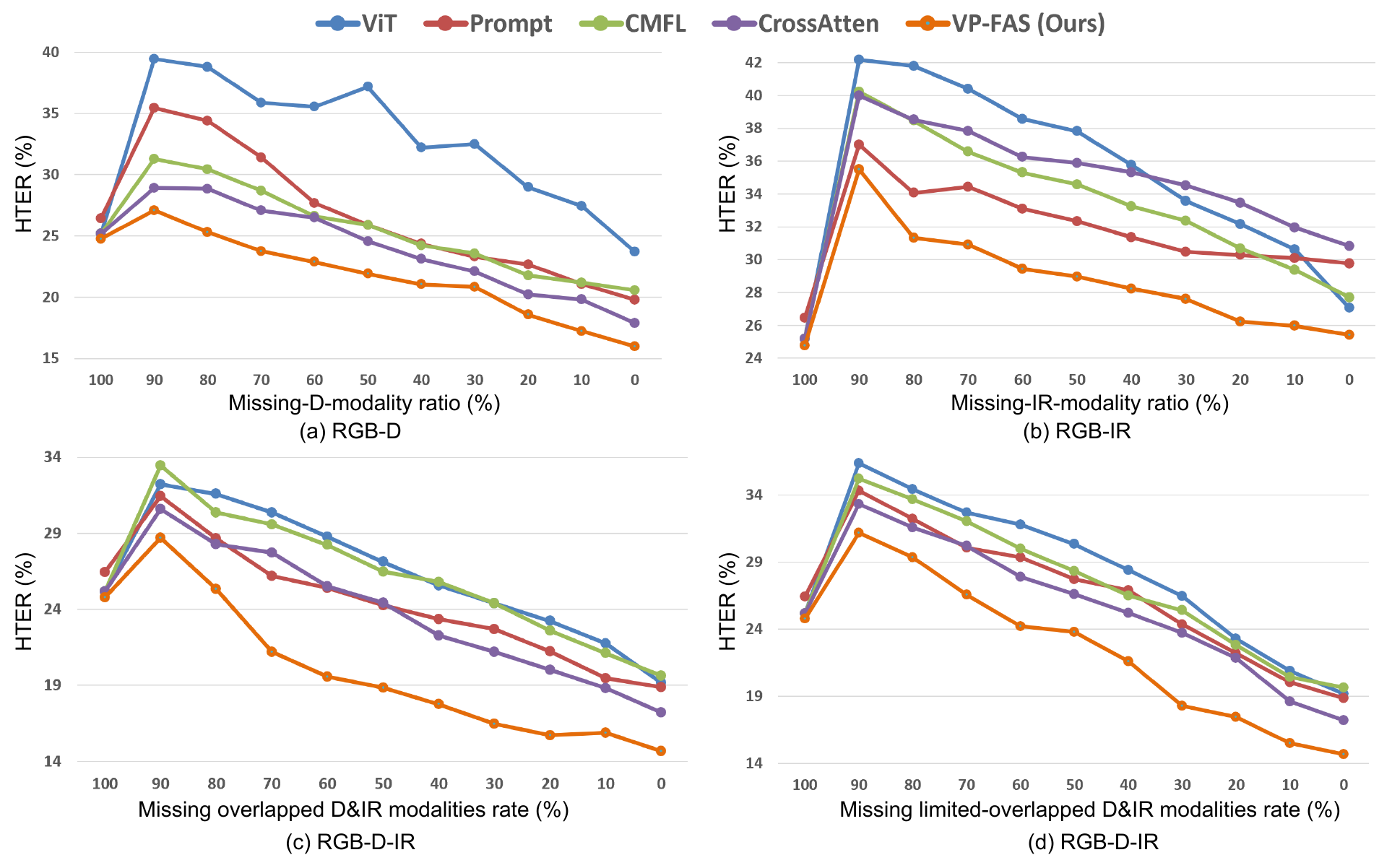}
\vspace{-0.8em}
  \caption{\small{
 Cross-testing results from WMCA to CASIA-SURF with different modality-missing ratios under four flexible-modal settings. }
  }
\label{fig:cross1}
\vspace{-0.3em}
\end{figure*}

\begin{figure*}
\centering
%\vspace{-0.5em}
\includegraphics[scale=0.41]{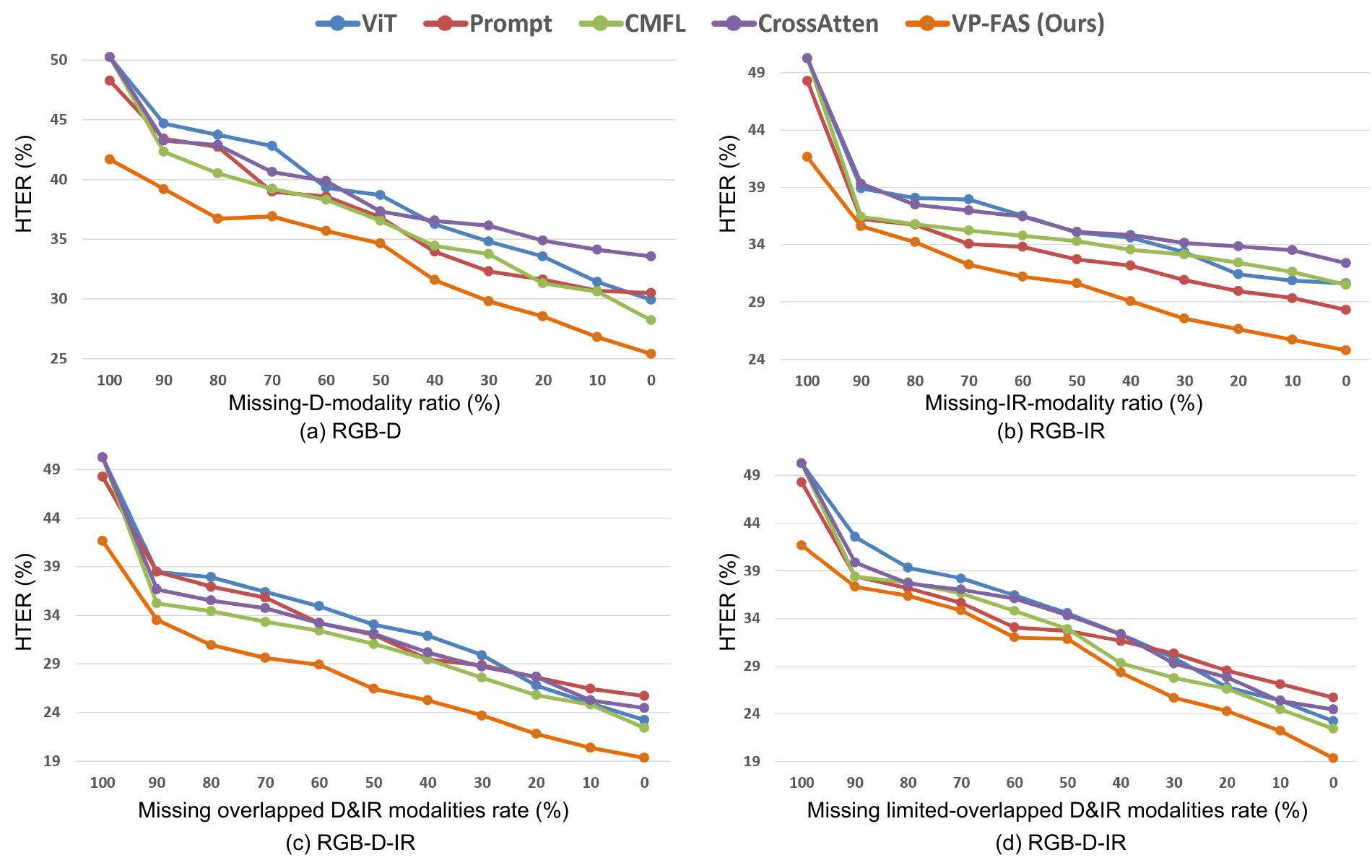}
\vspace{-0.8em}
  \caption{\small{
 Cross-testing results from CASIA-SURF to WMCA with different modality-missing ratios under four flexible-modal settings.}
  }
\label{fig:cross2}
\vspace{-0.3em}
\end{figure*}

\vspace{0.3em}
\noindent\textbf{Impact of missing-modality regularization (MMR).}\quad 
Here we discuss how the MMR influences the VP-FAS under flexible-modal RGB-D and RGB-IR settings. As shown in the last three rows of Table~\ref{tab:ablation}, compared with `VP-FAS w/o MMR', the proposed `VP-FAS' decreases 2.29\% and 0.45\% ACER under flexible-modal RGB-D and RGB-IR setting, respectively. In other words, MMR is able to force models to learn consistent multimodal feature embeddings when missing partial modalities, thus multimodal models trained with MMR are less sensitive to missing modalities. In Equation (5), the `sg' (stop gradient) operation is used in the right term, which can alleviate the conflicted information exchange during the flexible-modal model updating. We can also see from the result `VP-FAS w/o `sg' in MMR' that the performance degrades sharply when the stop gradient operation is missing.

\vspace{0.3em}
\noindent\textbf{Impact of $\theta$ and $d'$ in residual contextual prompt.}\quad 
Here we study the central difference intensity $\theta$ and hidden dimension $d'$ in the residual contextual prompt. As for the central difference intensity $\theta$, we find from Figure~\ref{fig:ablation}(a) that the smallest ACER values can be obtained when $\theta$=0.5 for both flexible-modal settings, indicating similar importance of local intensity and fine-grained gradient features can effectively guide the residual contextual prompt learning. In terms of hidden dimension $d'$, it can be seen from Figure~\ref{fig:ablation}(b) that despite being more lightweight, lower dimensions (16 and 32) cannot achieve satisfactory performance due to weak representation capacity. The best flexible-modal performance can be achieved when $D'$=64 in both RGB-D and RGB-IR settings.

\vspace{0.3em}
\noindent\textbf{Impact of prompt length $p$ in VP-FAS.}\quad
Figure~\ref{fig:ablation}(c) illustrates the flexible-modal performance with different prompt lengths $p$. Please note that in VP-FAS, defaulted combining setting with $p$//2 vanilla visual prompts and $p$//2 residual contextual prompts is utilized. It is clear that prompt length influences the flexible-modal RGB-D scenario a lot while making less impact on the flexible-modal RGB-IR scenario. The best flexible-modal performance can be achieved when $p$=40 while only introducing 3.6\% trainable parameters compared with finetuning whole ViT models.

\vspace{-1.3em}
\subsection{More Quantitative Results and Discussion}
\label{sec:quantitative}
We conduct further experiments to analyze the robustness of our proposed method against eleven equidistant different modality-missing ratios $\alpha$ between training and testing phases. Particularly, $\alpha$=100\% and $\alpha$=0\% indicate training and testing with only RGB unimodal data and complete multimodal data, respectively. 

\vspace{0.3em}
\noindent\textbf{Intra-testing robustness to different $\alpha$.} \quad   
We evaluate intra-testing performance on WMCA and CASIA-SURF with four flexible-modal settings. As shown in Figure~\ref{fig:intra1}, we find that on WMCA, 1) except fluctuation of the `Prompt' model with flexible-modal RGB-D setting, all other methods within four flexible-modal settings have a consistent trend of ACER decreasing when $\alpha$ becomes larger and larger; and 2) `CMFL' and the proposed `VP-FAS' are more robust to larger $\alpha$, compared with other three baseline methods. Meanwhile, the performance gaps of different methods are closer when $\alpha$ becomes smaller and smaller. Moreover, we can find similar conclusions on CASIA-SURF from Figure~\ref{fig:intra2} that the proposed `VP-FAS' is less sensitive to missing-modality scenarios. It is also interesting to find from Figures~\ref{fig:intra2}(c)(d) that on CASIA-SURF, performance gaps on two flexible RGB-D-IR settings are close, which might be caused by the issue of low-quality complete-modality data in CASIA-SURF.

\vspace{0.2em}
\noindent\textbf{Cross-testing robustness to different $\alpha$.} \quad   
Here we evaluate and discuss cross-testing generalization between WMCA and CASIA-SURF with four flexible-modal settings. Please note that it is challenging as the flexible-modal models should overcome the domain shift issue and missing-modality corruption together. As illustrated in Figure~\ref{fig:cross1}, we find that when trained on WMCA and tested on CASIA-SURF, compared with baseline `ViT', prompt learning based methods (see results from `Prompt' and `VP-FAS') are more generalized and robust to different $\alpha$ under all four flexible-modal settings. Similar trends when trained models on CASIA-SURF and tested on WMCA can also be observed in Figure~\ref{fig:cross2} that prompt learning based methods including `Prompt' and `VP-FAS' are less sensitive to high missing-modality scenarios (when $\alpha\geq$40\%).

\section{Conclusion and Future work}
%\vspace{-0.1em}
\label{sec:conc}

In this paper, we propose Visual Prompt flexible-modal Face Anti-Spoofing (VP-FAS) to plug both vanilla visual prompts and residual contextual prompts into multimodal transformers, which only requires 3.6\% trainable parameters compared with finetuning whole ViT models. Besides, we propose the missing-modality regularization to force models to learn robust flexible-modal features. We also introduce novel scenarios and evaluate classical baselines for flexible-modal FAS, where the missing modality may occur differently for each data sample, either in the training or testing phase. Extensive experiments on four novel flexible-modal settings demonstrate the effectiveness of the proposed method.

In the future, we will 1) investigate VP-FAS under more practical continual flexible-modal scenarios, i.e., sequential unimodal or multimodal data from agnostic domains; and 2) explore mobile-efficient VP-FAS models for mobile-friendly flexible-modal deployment.

\ifCLASSOPTIONcaptionsoff
  \newpage
\fi

\bibliographystyle{IEEEtran}
\bibliography{IEEEabrv,reference}

\end{document}